\documentclass[letterpaper, 10 pt, conference]{ieeeconf}  

\IEEEoverridecommandlockouts                              

\usepackage[T1]{fontenc}
\usepackage{graphics} 
\usepackage{epsfig} 
\usepackage{times} 
\usepackage{amsmath} 
\usepackage{amssymb}  
\usepackage{url}
\usepackage{threeparttable}
\usepackage{multirow}
\usepackage{makecell}
\usepackage{bbold}
\usepackage{bbm}
\usepackage{hyperref}
\usepackage{tikz}
\usetikzlibrary{calc}
\usetikzlibrary{shapes.geometric, arrows, decorations.pathreplacing}
\usetikzlibrary{backgrounds}
\usepackage{xcolor}
\usepackage{adjustbox}
\usepackage{fancyhdr}

\definecolor{network-blue}{RGB}{165, 192, 221}
\definecolor{light-yellow}{RGB}{238, 233, 218}
\definecolor{light-green}{RGB}{211, 235, 205}
\definecolor{light-red}{RGB}{242, 182, 160}
\definecolor{light-blue}{RGB}{124, 150, 171}
\definecolor{green}{RGB}{105,135,105}
\definecolor{red}{RGB}{231,97,97}
\definecolor{environment-c}{RGB}{254, 242, 244}
\definecolor{emb-c}{RGB}{0, 51, 124}
\definecolor{gt-c}{RGB}{155, 164, 181}
\definecolor{standard-gaussian}{RGB}{230, 226, 195}
\definecolor{prior-distribution}{RGB}{136, 164, 124}
\definecolor{skill-instance}{RGB}{28, 49, 94}

\title{\LARGE \bf
Penalty-Based Imitation Learning With Cross Semantics Generation Sensor Fusion for Autonomous Driving
}

\author{Hongkuan Zhou$^{2\, *}$, Aifen Sui $^{1\, \dagger}$, Letian Shi$^{2\, *}$, and Yinxian Li$^{2\,}$
\thanks{$^{1}$A. Sui  is at the Trustworthy Technology and Engineering Laboratory, Huawei Munich Research Center.}%
\thanks{$^{2}$H. Zhou, L. Shi and Y. Li are students at TUM School of Computation, Information and Technology, Technical University of Munich. This work was done during their internship at Huawei Munich Research Center.}%
\thanks{$^{\dagger}$Corresponding author: A. Sui {\tt\small aifen.sui@huawei.com}}
\thanks{$^{*}$Equally contribution.}
}

\begin{document}

\maketitle
\thispagestyle{fancy}
\lhead{Accepted by ITSC 2023}
\pagestyle{empty}


\begin{abstract}

In recent times, there has been a growing focus on end-to-end autonomous driving technologies. This technology involves the replacement of the entire driving pipeline with a single neural network, which has a simpler structure and faster inference time. However, while this approach reduces the number of components in the driving pipeline, it also presents challenges related to interpretability and safety. For instance, the trained policy may not always comply with traffic rules, and it is difficult to determine the reason for such misbehavior due to the lack of intermediate outputs. Additionally, the successful implementation of autonomous driving technology heavily depends on the reliable and expedient processing of sensory data to accurately perceive the surrounding environment. In this paper, we provide penalty-based imitation learning approach combined with cross semantics generation sensor fusion technologies (P-CSG) to efficiently integrate multiple modalities of information and enable the autonomous agent to effectively adhere to traffic regulations. Our model undergoes evaluation within the Town 05 Long benchmark, where we observe a remarkable increase in the driving score by more than 12\% when compared to the state-of-the-art (SOTA) model, InterFuser. Notably, our model achieves this performance enhancement while achieving a 7-fold increase in inference speed and reducing the model size by approximately 30\%. For more detailed information, including code-based resources, they can be found at \url{https://hk-zh.github.io/p-csg/}

\end{abstract}

\section{Introduction}
Autonomous driving is an emerging field of research at the intersection of robotics and computer vision. 
Recently, end-to-end autonomous driving \cite{7995975} \cite{chen2019lbc} \cite{Chitta2021ICCV(NEAT)}, integrating the perception module and decision-making module into one learning system to optimize, gains popularity in the researches as it proved surprisingly powerful with minimum training data gained from simulation environment.
However, end-to-end approach still suffers from the problem of interpretability and can not guarantee the most important factor ``safety'' in autonomous driving. Our primary objective is to enhance the safety of the end-to-end system through two main approaches. Firstly, we aim to enhance the reliability of the multi-sensor fusion algorithm, which will enable the system to perceive its surrounding environment with greater accuracy and robustness. Secondly, in order to enhance the interpretability of the end-to-end approach, we focus on refining the policy learning algorithm, enabling the autonomous agent to effectively adhere to traffic regulations.

The fusion of LiDAR and RGB sensors recently show impressive results in the context of autonomous driving. LiDAR sensors provide accurate 3D information of surrounding environment while they lack color information compared to RGB sensors; RGB sensors are more suitable to recognize traffic lights and traffic sign patterns while they are not resilient to bright light and other bad weather conditions compared to LiDAR sensors. Some fusion technologies \cite{deep-continuous-fusion} \cite{liu2022bevfusion} have achieved commendable results in the field of object detection. In terms of end-to-end autonomous driving, \cite{TransFuser} \cite{TransFuser+} \cite{shao2022interfuser} focus more on attention-based approaches to extract the global context from different modalities. Despite its potential, the additional Transformer architecture leads to a significant increase in both the training time and inference time of the model. To address this issue, we fuse the information (Figure \ref{fig:Shared semantic features}) obtained from LiDAR and RGB by aligning their shared semantic information 
with auxiliary losses. This approach requires fewer parameters, yet it still produces remarkable results. 


\tikzstyle{legend} = [rectangle, minimum width=0.5cm, align=center, rounded corners, opacity=0.75]
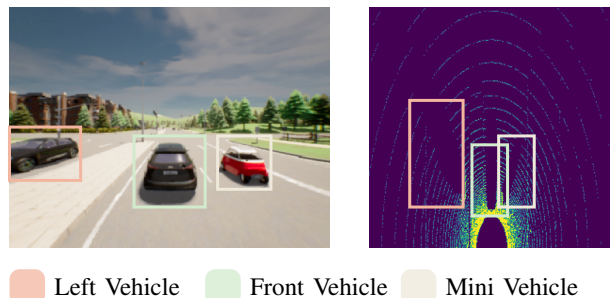
\begin{figure}
    \centering
    \begin{adjustbox}{width=0.45\textwidth}
        \begin{tikzpicture}[node distance=2cm]
        \node[inner sep=0pt] (rgb-image) at (0,0)
        {\includegraphics[width=4.5cm]{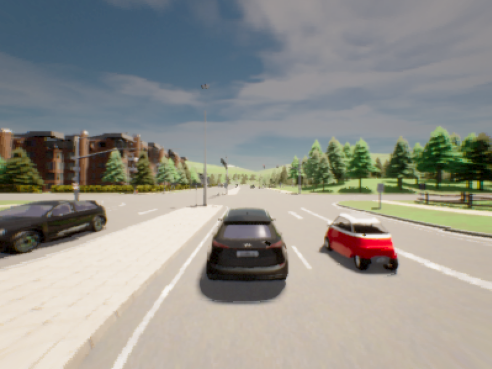}};
        \node[inner sep=0pt, right of = rgb-image, xshift=2.5cm] (lidar-image)
        {\includegraphics[width=3.375cm]{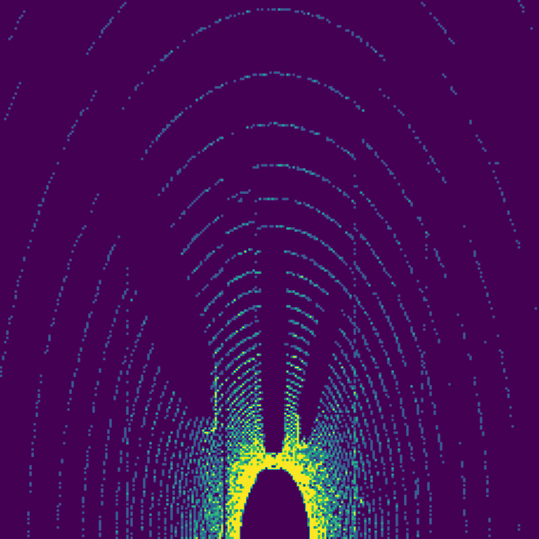}};
        \node[rectangle, left of = rgb-image, draw=light-red, minimum height=0.75cm, minimum width=1cm, xshift=0.25cm, yshift=-0.375cm, very thick](w1){};
        \node[rectangle, below of = rgb-image, draw=light-green, minimum height=1cm, minimum width=1cm, xshift=0cm, yshift=1.375cm, very thick](w2){};
        \node[rectangle, right of = rgb-image, draw=light-yellow, minimum height=0.75cm, minimum width=0.75cm, xshift=-0.95cm, yshift=-0.5cm, very thick](w3){};
        
        \node[rectangle, left of = lidar-image, draw=light-red, minimum height=1.5cm, minimum width=0.75cm, xshift=1.25cm, yshift=-0.375cm, very thick](w1){};
        \node[rectangle, below of = lidar-image, draw=light-green, minimum height=1cm, minimum width=0.5cm, xshift=0cm, yshift=1.25cm, very thick](w2){};
        \node[rectangle, right of = lidar-image, draw=light-yellow, minimum height=1cm, minimum width=0.5cm, xshift=-1.625cm, yshift=-0.625cm, very thick](w3){};
        
        \node (legend1) [legend, below of=rgb-image, xshift=-2cm, yshift=-0.25cm, minimum width=0.5cm, minimum height=0.5cm, fill=light-red, label={[label distance=0cm]0: Left Vehicle}]  {};
        \node (legend2) [legend, right of=legend1, xshift=0.75cm, minimum width=0.5cm, minimum height=0.5cm, fill=light-green, label={[label distance=0cm]0: Front Vehicle}]  {};
        \node (legend3) [legend, right of=legend2, xshift=0.75cm, minimum width=0.5cm, minimum height=0.5cm, fill=light-yellow, label={[label distance=0cm]0: Mini Vehicle}]  {};
        \end{tikzpicture}
 	\end{adjustbox}
 	\caption{\textbf{Illustration.} To safely navigate in the road, the ego-vehicle must capture the surrounding context from the RGB camera (left) and  LiDAR (right). Our P-CSG model integrates both modalities by capturing shared semantic features via feature alignment and cross semantic generation. }
 	\label{fig:Shared semantic features}

\end{figure}
Reinforcement learning (RL) \cite{sutton1998introduction} and imitation learning (IL) \cite{Imitation_Learning} are two main learning paradigms used by end-to-end autonomous driving. Even though reinforcement learning demonstrates huge potential in autonomous driving, it often confronts limitations due to low sample efficiency and the requirement for careful reward development to learn, which poses challenges in obtaining sufficient training data for effective learning. Meanwhile, RL algorithms may also learn to take risky actions that lead to accidents or unsafe driving behaviors. Consequently, other researchers have turned to imitation learning approaches. However, current imitation learning approaches still lack effective mechanism to ensure safety during training process. After careful study, we found that the metric of autonomous driving and the objective function of imitation learning are not unified which means a low loss of learning objective does not guarantee the good performance of the agent in the testing environment. The traffic rules e.g. forbidding running a red light and stop sign are not reflected in the objective function. Our objective is to develop a novel objective function by incorporating penalty mechanisms, with the intention of augmenting the trained model's responsiveness to traffic rule violations. This integration aims to instill a heightened awareness of traffic regulations during the training process, ultimately leading to improved overall performance of the model.


Our main contributions to this paper can be summarized as follows:
\begin{itemize}
    \item We proposed a penalty-based imitation learning approach that leverages constraint optimizations to make the end-to-end autonomous driving model more sensitive to traffic rule violations. This objective function design also unifies the metric of autonomous driving and the objective of imitation learning.
    \item We proposed a novel multi-sensor fusion model to extract the shared features and unique features between different modalities, making it easier for the decision network to get a global context for policy generation.
\end{itemize}
Our suggested approach is termed as \textbf{P}enalty-based Imitation Learning coupled with \textbf{C}ross \textbf{S}emantics \textbf{G}eneration sensor fusion (P-CSG).

\section{related works}
\subsection{End-to-End Autonomous Driving}
Today's autonomous driving technologies have two main branches, modular and end-to-end approaches. Modular approaches apply a fine-grained pipeline of software modules working together to control the vehicle. In contrast, the entire pipeline of end-to-end driving is treated as one single learning task. End-to-end approaches have shown great success in computer vision tasks, such as object detection \cite{7112511} \cite{Fast_RCNN} \cite{Faster_RCNN} \cite{redmon2016you}, object tracking \cite{brasó2020learning}, and semantic segmentation \cite{ronneberger2015u} \cite{DBLP:journals/corr/CicekALBR16}. The success of these tasks builds a solid foundation for end-to-end autonomous driving. It is reasonable to believe end-to-end approaches are capable of solving autonomous driving problems in the near future. 
The most common learning methods for end-to-end autonomous driving are imitation learning \cite{TransFuser} \cite{latefusion} \cite{Filos2020CanAV} \cite{Chitta2021ICCV(NEAT)} \cite{chen2022lav} \cite{8813900} and reinforcement learning  \cite{DBLP:journals/corr/abs-2001-08726} \cite{CPRL}.

\subsection{Safety Mechanism in End-to-End Autonomous Driving}
In the realm of autonomous driving, a key challenge is implementing safety mechanisms that can prevent accidents and protect passengers, pedestrians, and other road users. Within the framework of imitation learning, the agent learns driving skills by emulating expert demonstrations. The quality of these demonstrations has a significant impact on the agent's ability to drive safely in traffic. To improve the safety of the autonomous driving agent, researchers in \cite{TransFuser+} focus on enhancing the quality of the expert agent, while those in \cite{shao2022interfuser} introduce an additional safety module that filters out potentially dangerous driving behaviors generated by the network. Our contribution is to introduce the ``Penalty'' concept to the imitation learning framework, which incentivizes the trained agent to adopt safer driving behaviors.

\subsection{Multi-sensor Fusion Technologies}
Sensor Fusion technologies are commonly employed for 3D object detection and motion forecasting. Among the various types of sensors that can be integrated, the fusion of LiDAR and camera sensors is most frequently employed, where LiDAR data serves as a supplement to image data, providing additional information about the surrounding environment and improving data reliability due to its consistency in various environments. There are three branches of sensor fusion: early fusion, middle fusion, and late fusion. In early fusion, the data is fused before being fed into the learnable system, which is the most efficient approach. In middle fusion, the information is merged in the middle of the network, and the fused features are used to produce task-specific outputs. Late fusion is an ensemble learning method that combines the outputs generated by each modality into a final result. 

Multi-sensor fusion has received much research attention in the field of end-to-end autonomous driving. Prior works such as LateFusion\cite{latefusion} used a large Multi-Layer Perception (MLP) network to process the features extracted by the perception networks of LiDAR and RGB inputs. This MLP layer takes the tasks of features weighting, selection, and fusion which makes it hard to 
capture a global context of multi-modality inputs. TransFuser\cite{TransFuser} provides an approach that leverages the attention mechanism to fuse the LiDAR and RGB information. They used the transformer architecture to achieve the multi-modality global context. The Transformer-based fusion model is applied to different resolutions between the LiDAR and RGB perception networks. TransFuser+ \cite{TransFuser+}, as an extension of TransFuser, introduced more headers in the neural networks which incorporate four auxiliary tasks: depth prediction and semantic segmentation from the image branch; HD map prediction, and vehicle object detection from the BEV branch. These auxiliary tasks help to visualize the black box of the whole network. In addition, this approach also guarantees important information flow in the latent space because the information contained in the latent space should not only be able to complete the navigation task but also manually pre-defined auxiliary tasks.

\section{Methodologies}

In this section, we propose a novel multi-sensor fusion approach and a penalty-based Imitation Learning paradigm for end-to-end autonomous driving.
\subsection{Problem Setting}
The task we concentrate on is point-to-point navigation in an urban setting where the goal is to complete a route with safe reactions to dynamic agents such as moving vehicles and pedestrians. The traffic rules should also be followed. 

\textbf{Imitation Learning (IL):} Imitation Learning can learn a policy $\pi$ that clone the behavior of an expert policy $\pi^*$. In our setup, the policy is conditioned on the multi-modalities inputs of current observations. We used the Behavior Clone (BC) approach of IL. An expert policy is applied in the environment to collect a large dataset $\mathcal{D}=\{(\textbf{x}^i, \textbf{w}^i)\}_{i=1}^{Z}$ with the size of $Z$, which contains the observation of the environment $\textbf{x}^i$ and a set of waypoints $\textbf{w}^i$ in the future timesteps. The objective function is defined as:
\begin{equation}
    \label{eq:objective function}
    \mathcal{F} = \mathbb{E}_{(\mathcal{X},\mathcal{W})\sim \mathcal{D}}[\mathcal{L}(\mathcal{W}, \pi(\mathcal{X}))]
\end{equation}
where $\mathcal{L}$ is the loss function. 

In our setting, the observation $\mathcal{X}$ consists of one RGB image and one LiDAR point cloud from the current time step. We used only one single frame since other works since \cite{DBLP:journals/corr/abs-1905-06937}, \cite{DBLP:journals/corr/abs-1812-03079} have shown that using multiple frames does not improve the information gain much. A PID controller $\mathcal{I}$ is applied to perform low-level control, i.e. steer, throttle, and brake based on these predicted future waypoints. 

\textbf{Global Planner:} According to CARLA \cite{Dosovitskiy17} 0.9.10's protocol, the high-level goal locations $G$ is provided as GPS coordinates. This goal location $G$ is sparse (hundreds of meters apart) which can only be used as guidance. In contrast, those to be predicted waypoints are dense, only a few meters way away from each other. 

\tikzstyle{module} = [rectangle, rounded corners, text centered, draw=black, fill=network-blue!70]
\tikzstyle{arrow} = [thick,->,>=stealth]
\tikzstyle{emb} = [rectangle, minimum width=0.5cm, align=center, draw=black, fill=emb-c!60, font=\scriptsize\linespread{0.9}\selectfont]
\tikzstyle{loss} = [rectangle, minimum width=0.5cm, align=center, draw=black, fill=white, font=\scriptsize\linespread{0.9}\selectfont]
\tikzstyle{dash-box} = [rectangle, dashed, thick, draw=black]

\begin{figure*}
    \centering
    \begin{adjustbox}{width=0.95\textwidth}
    \begin{tikzpicture}[node distance=2cm]
        \node[inner sep=0pt] (front) at (0,0)
        {\includegraphics[width=2.5cm]{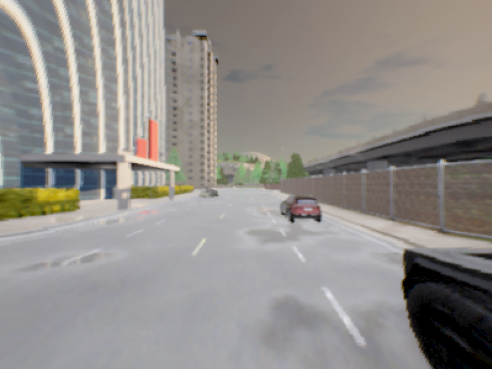}};
        \node[inner sep=0pt, below of=front, yshift=-3cm, xshift=-0.25cm] (lidar2)
        {\includegraphics[width=2cm]{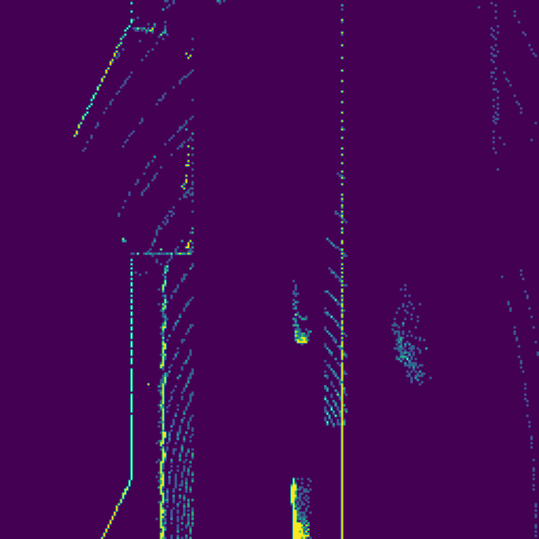}};
        \node[inner sep=0pt, below of=front, yshift=-3.5cm, xshift=0.25cm] (lidar1)
        {\includegraphics[width=2cm]{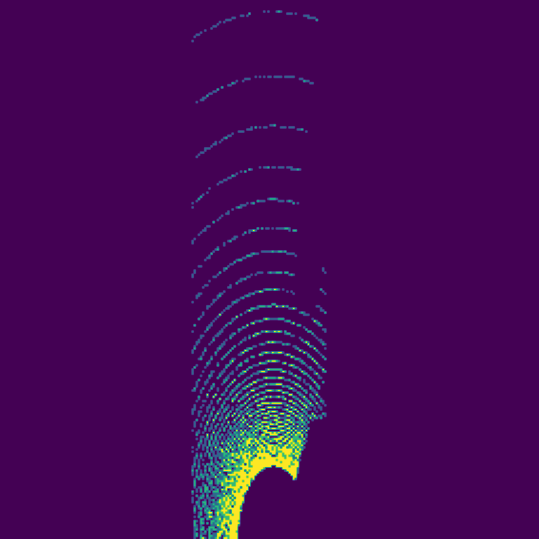}};
        
        \node (legend1) [emb, below of=lidar1, xshift=-1.25cm, yshift=-0.25cm, minimum width=0.5cm, minimum height=0.5cm, fill=light-yellow, label={[label distance=0cm]0: Shared Features}]  {};
        \node (legend2) [emb, right of=legend1, xshift=1cm, minimum width=0.5cm, minimum height=0.5cm, fill=light-green, label={[label distance=0cm]0: Unique Features}]  {};
        \node (legend3) [emb, right of=legend2, xshift=1.1cm, minimum width=0.5cm, minimum height=0.5cm, fill=light-red, label={[label distance=0cm]0: Measurements}]  {};
        
        \node (resnet34) [module,  right of=front, text width=2cm, align=center, xshift=1cm, minimum height=1.5cm] {Resnet34};
        \node (resnet18) [module,  right of=lidar1, text width=2cm, align=center, xshift=0.75cm, yshift=0.25cm, minimum height=1.5cm] {Resnet18};
        \node (rgb-emb) [emb, right of=resnet34, xshift=0.25, minimum height=1.5cm] { 5\\1\\2};
        \node (lidar-emb) [emb, right of=resnet18, xshift=0.25, minimum height=1.5cm] {5\\1\\2};
        \node (mlp1) [module,  right of=rgb-emb, text width=1cm, align=center, xshift=0.25cm, yshift=1cm, minimum height=1.5cm] {MLP};
        \node (mlp2) [module,  right of=rgb-emb, text width=1cm, align=center, xshift=0.25cm, yshift=-1cm, minimum height=1.5cm] {MLP};
        \node (align-loss) [loss, below of=mlp2, yshift=0.375cm, minimum height=0.75cm, minimum width=1.25cm] {Align.\\Loss};
        \node (mlp3) [module,  right of=lidar-emb, text width=1cm, align=center, xshift=0.25cm, yshift=1cm, minimum height=1.5cm] {MLP};
        \node (mlp4) [module,  right of=lidar-emb, text width=1cm, align=center, xshift=0.25cm, yshift=-1cm, minimum height=1.5cm] {MLP};
        \node (rgb-emb-u) [emb, right of=mlp1, xshift=-0.25cm, minimum height=1.5cm, fill=light-green] {1\\2\\8};
        \node (rgb-emb-s) [emb, right of=mlp2, xshift=-0.25cm, minimum height=1.5cm, fill=light-yellow] {1\\2\\8};
        \node (lidar-emb-s) [emb, right of=mlp3, xshift=-0.25cm, minimum height=1.5cm, fill=light-yellow] {1\\2\\8};
        \node (lidar-emb-u) [emb, right of=mlp4, xshift=-0.25cm, minimum height=1.5cm, fill=light-green] {1\\2\\8};
        \node (measurements) [emb, below of=lidar-emb-u, yshift=0.5cm, minimum height=0.5cm, fill=light-red] {4};
        \node (dash-box1) [dash-box, right of=mlp2, xshift=-0.25cm, yshift=-2.125cm, minimum width=1cm, minimum height=10.25cm, draw=emb-c!60]{};
        
        \node (decoder1) [module,  right of=rgb-emb-u, text width=1.25cm, align=center, xshift=1cm, minimum height=1.5cm] {Decoder};
        \node (decoder2) [module,  right of=rgb-emb-s, text width=1.25cm, align=center, xshift=1cm, yshift=-0.5cm, minimum height=1.5cm] {Decoder};
        \node (decoder3) [module,  right of=lidar-emb-s, text width=1.25cm, align=center, xshift=1cm, yshift=0.5cm, minimum height=1.5cm] {Decoder};
        \node (mlp5) [module,  right of=lidar-emb-u, text width=1cm, align=center, xshift=-0.25cm, minimum height=1.5cm] {MLP};
        \node (emb-c) [emb, right of=mlp5, xshift=-0.5cm, minimum height=1.5cm] {6\\4};

        \node (gru-decoder) [module,  right of=emb-c, text width=1.25cm, align=center, xshift=-0.375cm, minimum height=1.5cm] {GRU\\Decoder};
        \node (goal-location) [emb, below of=gru-decoder, minimum height=0.75cm, yshift=0.5cm] {Goal\\Location};

        \node (dash-box2) [dash-box, right of=decoder1, xshift=1.5cm, yshift=0cm, minimum width=2.5cm, minimum height=1.5cm]{};
        \node[inner sep=0pt, left of=dash-box2, yshift=0cm, xshift=1.375cm] (stopsign){\includegraphics[width=1cm]{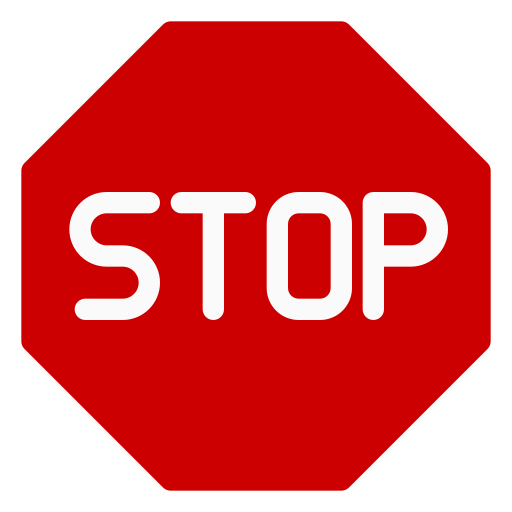}};
        \node[inner sep=0pt, right of=stopsign, yshift=0cm, xshift=-0.75cm] (stopsign){\includegraphics[width=1cm]{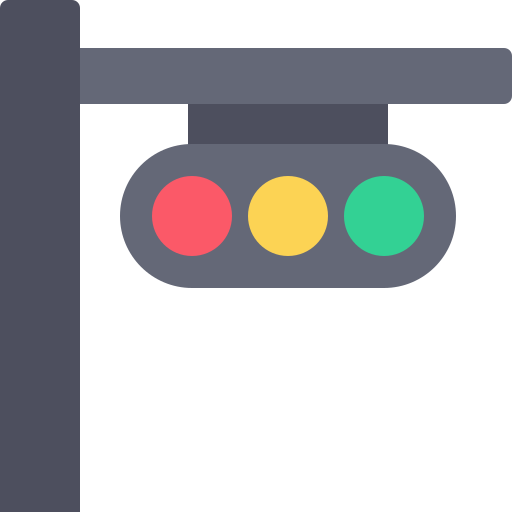}};
        
        \node[inner sep=0pt, right of=decoder2, xshift=1.5cm] (front-seg){\includegraphics[width=2.5cm]{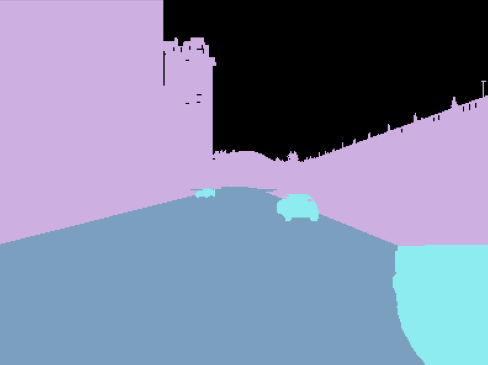}};
        \node[inner sep=0pt, right of=decoder3, xshift=1.75cm] (topdown-seg){\includegraphics[width=2cm]{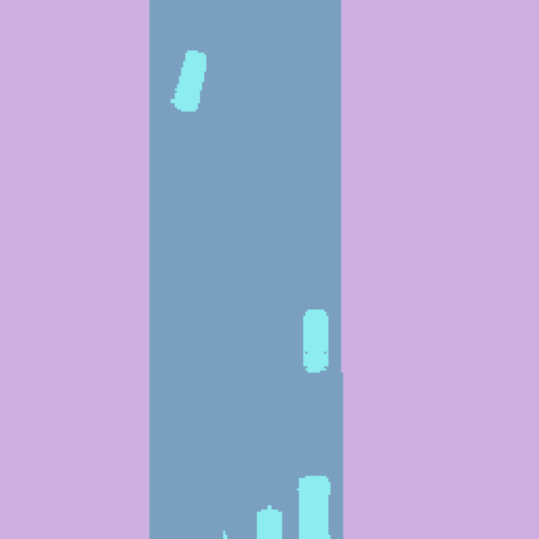}};
        \node (waypoints) [emb, right of=gru-decoder, minimum height=0.75cm, minimum width=1.5cm] {Waypoints};
        \draw [arrow] (front.east) -- (front.east-|resnet34.west);
        \draw [arrow] (lidar1.east|-resnet18.west) -- (resnet18.west);
        \draw [arrow] (lidar1.east|-resnet18.west) -- (resnet18.west);
        \draw [arrow] (resnet34.east) -- (resnet34.east -| rgb-emb.west);
        \draw [arrow] (resnet18.east) -- (resnet18.east -| lidar-emb.west);
        \draw [arrow] (rgb-emb.east) -- +(+0.625, 0) |- (mlp1.west);
        \draw [arrow] (rgb-emb.east) -- +(+0.625, 0) |- (mlp2.west);
        \draw [arrow] (lidar-emb.east) -- +(+0.625, 0) |- (mlp3.west);
        \draw [arrow] (lidar-emb.east) -- +(+0.625, 0) |- (mlp4.west);
        \draw [arrow] (mlp1.east) -- (mlp1.east-|rgb-emb-u.west);
        \draw [arrow] (mlp2.east) -- (mlp2.east-|rgb-emb-s.west);
        \draw [arrow] (mlp3.east) -- (mlp3.east-|lidar-emb-s.west);
        \draw [arrow] (mlp4.east) -- (mlp4.east-|lidar-emb-u.west);
        \draw [arrow] (rgb-emb-u.east) -- (rgb-emb-u.east-|decoder1.west);
        \draw [arrow] (rgb-emb-s.east) -- +(+0.25, 0) -- ($ (decoder3.west)+(-0.25, 0) $) -- (decoder3.west);
        \draw [arrow] (lidar-emb-s.east) -- +(+0.25, 0) -- ($ (decoder2.west)+(-0.25, 0) $) -- (decoder2.west);
        \draw [arrow, dashed, draw=emb-c!60] (dash-box1.east |- mlp5.west) -- (mlp5.west);
        \draw [arrow] (mlp5.east) -- (emb-c.west);
        \draw [arrow] (emb-c.east) -- (gru-decoder.west);
        \draw [arrow] (gru-decoder.east) -- (waypoints.west);
        \draw [arrow] (goal-location.north) -- (gru-decoder.south);
        \draw [arrow] (decoder1.east) -- (dash-box2.west);
        \draw [arrow] (decoder2.east) -- (front-seg.west);
        \draw [arrow] (decoder3.east) -- (topdown-seg.west);

        \draw [arrow] (rgb-emb-s.south) |- (align-loss.east);
        \draw [arrow] (lidar-emb-s.north) |- (align-loss.east);
    \end{tikzpicture}
    \end{adjustbox}
    \caption{\textbf{Architecture.} The top-down LiDAR pseudo image and front camera image go through two residual networks to extract 512 dimension feature vectors. We use four different MLPs to extract the shared features and the unique features. The unique features of RGB input are used to generate stop signs and traffic light indicators. The shared features of LiDAR are used to reconstruct the segmentation of RGB input while the shared features of RGB are used to reconstruct the segmentation of top-down LiDAR input. An alignment loss is used to align the shared features from LiDAR and camera inputs into the same space. These shared features and unique features are concatenated along with the measurements (velocity, throttle, steer, brake from the last frame) and then go through one MLP to reduce the size. Finally, they will be fed into one GRU decoder to predict short-term waypoints.}
    \label{fig:Architecture}
\end{figure*}
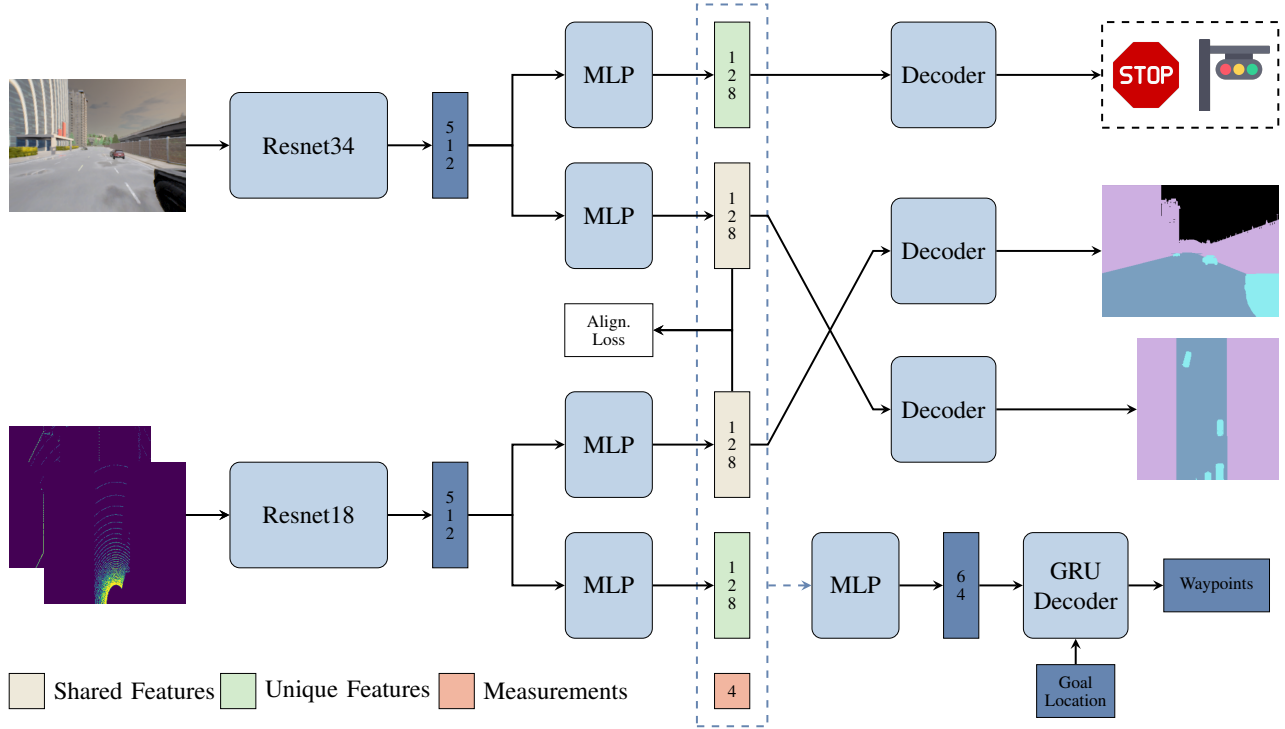

\subsection{Cross Semantics Generation}

The motivation of our approach is based on the fact that multi-modal inputs have shared semantic information and also unique information. For instance, the geometric attributes and spatial coordinates of both vehicles and pedestrians are shared information that can be extracted from both LiDAR and RGB input. Figure \ref{fig:Shared semantic features} demonstrates the shared information of LiDAR and RGB input. The unique information refers to the complementary information that other inputs do not have. In the case of RGB input, unique information often pertains to features such as the color of traffic lights, patterns on traffic signs, and similar attributes. On the other hand, in the context of LiDAR input, unique information pertains to spatial relationships of objects. Our multi-sensor fusion approach aims to extract and align the shared features from LiDAR and RGB input sources so that the later decision network can leverage the organized features to achieve better performance.

To extract the shared information from LiDAR and RGB inputs, we propose cross semantics generation sensor fusion. As Figure \ref{fig:Architecture} demonstrates, the front RGB and top-down pre-processed LiDAR pseudo images will first be fed into two residual networks \cite{DBLP:journals/corr/HeZRS15} to extract the corresponding RGB and LiDAR features. Note that the LiDAR point cloud is pre-processed into the bird's eye view pseudo images which is the same setting as \cite{TransFuser}. We use four different linear layers to extract the shared features and unique features of LiDAR and RGB. The shared features of  RGB are used to generate the top-down semantic segmentation align with LiDAR input; The shared features of LiDAR are used to generate the semantic segmentation of corresponding RGB input. We refer this approach as cross semantics generation since the information from one modality is utilized to generate semantic representations of the other modality. In this way, the information flow is said to be 'crossed', as each modality contributes to the understanding of the other. The extracted shared features will maximized since the information derived from one modality should strive to generate an accurate semantic segmentation of the other modality to the best of its ability. An extra L2 loss is introduced to align the shared features of RGB and LiDAR into the same latent space. In our setup, the semantic segmentation contains 4 channels, the drivable area, the non-drivable area, the object in the drivable areas like vehicles and pedestrians, and others. In terms of the unique features from RGB input, we mainly concentrate on the traffic lights and stop signs. As we can see from the figure, the unique features from RGB input are used to train the traffic light and stop sign indicator which ensures the important information flows of traffic lights and stop signs in the neural network. These headers are also critical for later penalty-based Imitation which we will discuss in the following sections.

\subsection{Waypoint Prediction Network}
As shown in Figure \ref{fig:Architecture}, all the unique and shared features are concatenated into a 512-dimensional feature vector. This vector is fed into an MLP to reduce the dimension to 64 for computational efficiency reasons. The hidden layer of GRU is initialized with a 64-dimensional feature vector. GRU’s update gate controls the information flow from the hidden layer to the output. In each timestep, it also takes the current location and goal location as input. We follow the approach of \cite{Filos2020CanAV} that a single GRU layer is followed by a linear layer which takes the state of the hidden layer and predicts the relative position of the waypoint compared to the previous waypoint for $T=4$ time-steps. Hence, the predicted future waypoints are formed as $\{w_t = w_{t-1} + \delta w_t\}_{t=1}^T$. The start symbol for GRU is given by (0,0). 

\textbf{Controller}: Based on the predicted waypoints, we use two PID controllers for lateral and longitudinal directions respectively. We follow the settings of \cite{chen2019lbc}.

\subsection{Loss Functions}
Similar to previous works \cite{TransFuser}, \cite{chen2019lbc}, we also use $L_1$ loss as our reconstruction loss. For each input, the loss function can be formalized as:
\begin{equation}
    \mathcal{L} = \sum_{t=1}^{T} ||w_t - w_t^{gt}||_1
\end{equation}
where $w_t$ is the t-th predicted waypoints and $w_t^{gt}$ is the t-th ground truth waypoint produced by the expert policy. 

\textbf{Auxiliary Tasks}: In our cross semantics generation approach, we have four extra auxiliary tasks along with the main imitation learning task. As we explained in the above section, two of the auxiliary tasks are semantic segmentation. In order to ensure some important information flow in the network, we introduce two extra classification headers, namely traffic light classification and stop sign classification. These two headers help the neural network to capture traffic light and stop sign information which is significant for later penalty-based Imitation learning.

\textbf{Front View Semantics}. Front-view semantic segmentation has four different channels. We define $y_f$ as the ground truth 3D tensor with the dimension $H_f \times W_f \times 4$ and $\hat{y}_f$ as the output of the front view decoder with the same shape.  

\textbf{Top-down View Semantics}. Like front-view semantic segmentation, top-down-view semantic segmentation also has four channels. We define $y_{td}$ as the ground truth 3D tensor with the dimension $H_{td} \times H_{td} \times 4$ and $\hat{y}_{td}$ as the output of the top-down view decoder with the same shape. 

\textbf{Image-LiDAR Alignment Loss}. This loss aims to align the shared semantic features of Image and LiDAR into the same latent space. We use an L2-loss to align these features. 

\textbf{Traffic Light Classification}. The output of the traffic light decoder should be a vector of 4 which indicates these four states red light, yellow light, green light, and none in the current frame. We then define $y_l$ as the ground truth traffic light vector of length 4 and $\hat{y}_l$ as the output of the traffic light decoder with the same shape. 

\textbf{Stop Sign Classification}. The output of the stop sign decoder should have a vector of 2 which indicates if a stop sign exists in the current frame. The ground truth stop sign vector of length 2 and the output of the stop sign decoder with the same shape are defined as $y_s$ and $\hat{y}_s$, respectively. 
Based on what we defined above, the new loss function is given by:
\begin{equation}
\begin{aligned}
        \mathcal{L} = & \sum_{t=1}^{T} ||w_t - w_t^{gt}||_1 + \omega_f \mathcal{L}_{\rm CE}(y_{f},\hat{y}_{f}) + \\
        & \omega_{td} \mathcal{L}_{\rm CE}(y_{td},\hat{y}_{td}) + \omega_{l} \mathcal{L}_{\rm CE}(y_{l}, \hat{y}_{l}) +  \\ 
        & \omega_{s} \mathcal{L}_{\rm CE}(y_{s}, \hat{y}_{s}) + \omega_{a} \mathcal{L}_{2}(y_{s}, \hat{y}_{s}) 
\end{aligned}
\label{eq:new-designed objective}
\end{equation}
 where $\mathcal{L}_{CE}$ and $\mathcal{L}_2$ are the cross entropy loss and L2 loss, respectively. $\omega_f$, $\omega_{td}$, $\omega_{l}$,  $\omega_s$, $\omega_a$ are the weights for these auxiliary losses.

\subsection{Penalty-based Imitation Learning}
\tikzstyle{waypoint} = [ellipse, x radius=6pt, y radius=3pt]
\tikzstyle{legend} = [rectangle, minimum width=0.5cm, align=center, rounded corners, opacity=0.75]
\begin{figure*}
    \centering
    \begin{adjustbox}{width=0.95\textwidth}
        \begin{tikzpicture}[node distance=2cm]
            \node[inner sep=0pt, label={[label distance=0cm]90:Red Light Penalty}] (red-light-penalty1) at (0,0)
            {\includegraphics[width=5cm]{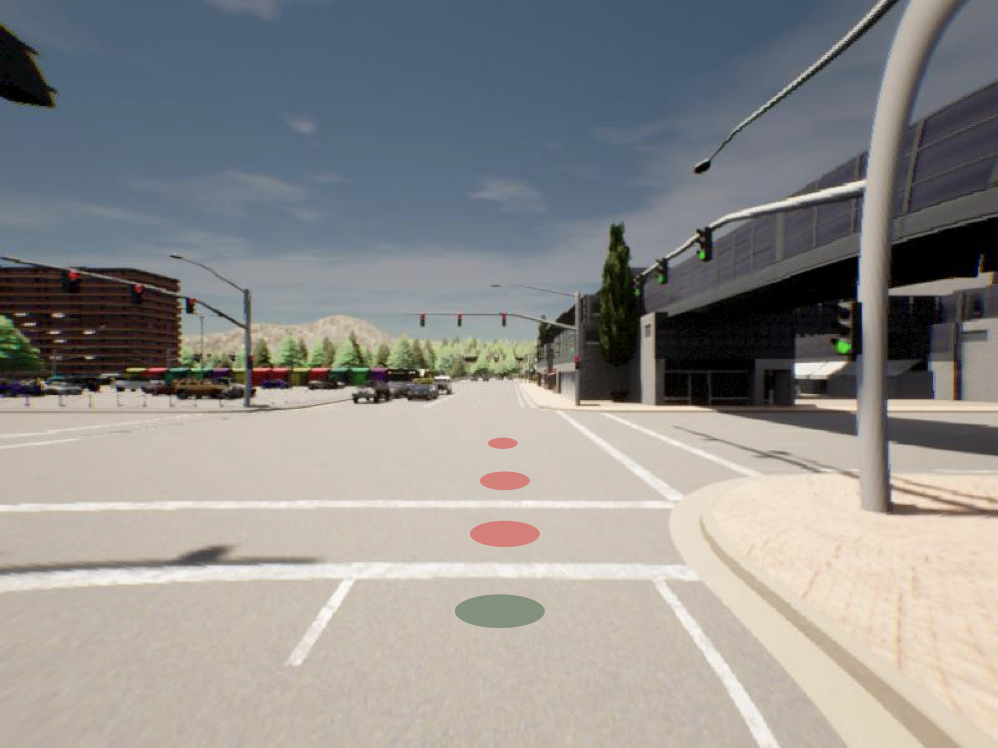}};
            \node[inner sep=0pt, below of=red-light-penalty1, yshift=-2cm] (red-light-penalty2)
            {\includegraphics[width=5cm]{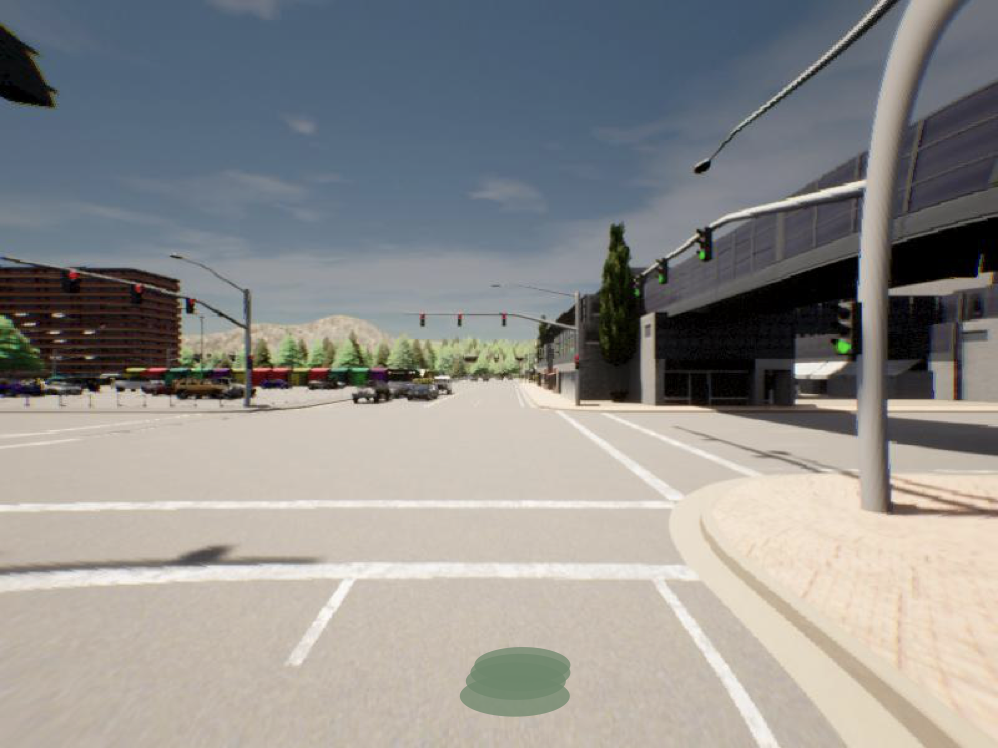}};
            
            \node[inner sep=0pt, right of= red-light-penalty1, xshift=3.5cm, label={[label distance=0cm]90:Stop Sign Penalty}] (stop-sign-penalty1) 
            {\includegraphics[width=5cm]{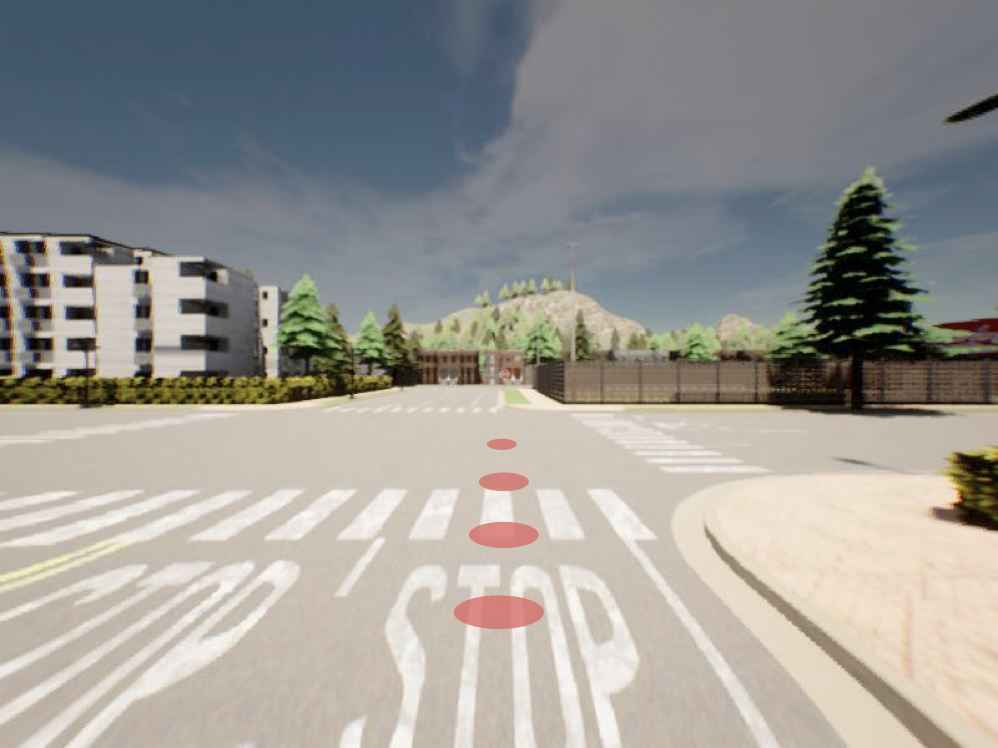}};
            \node[inner sep=0pt, below of=stop-sign-penalty1, yshift=-2cm] (stop-sign-penalty2)
            {\includegraphics[width=5cm]{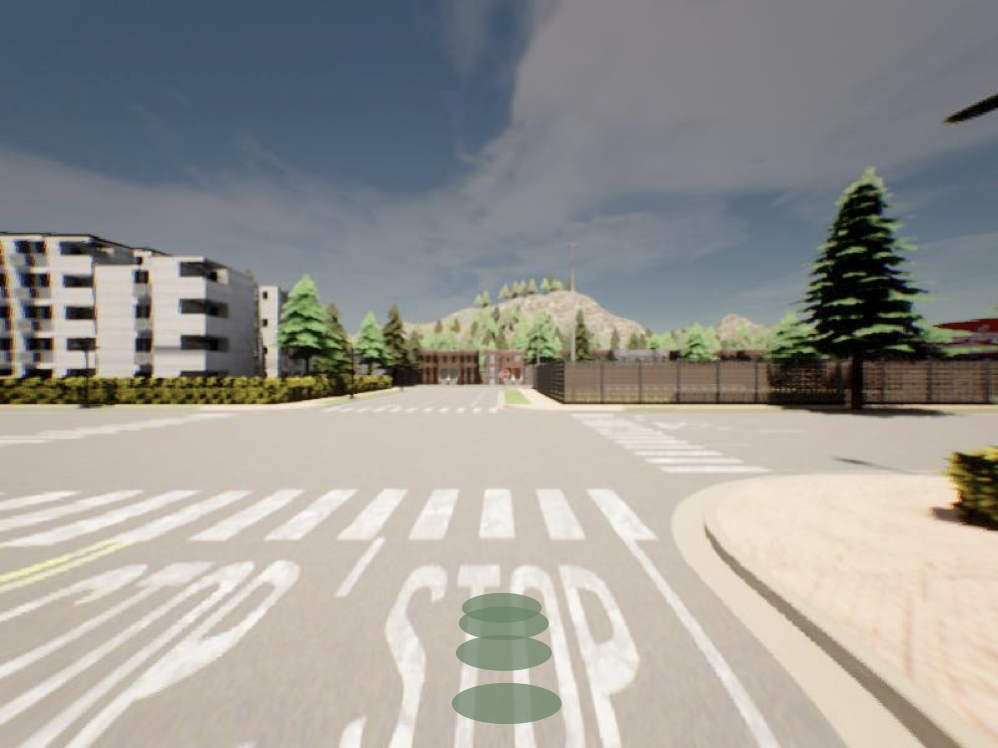}};
            
            \node[inner sep=0pt, right of= stop-sign-penalty1, xshift=3.5cm, label={[label distance=0cm]90:Speed  Penalty}] (speed-penalty1) 
            {\includegraphics[width=5cm]{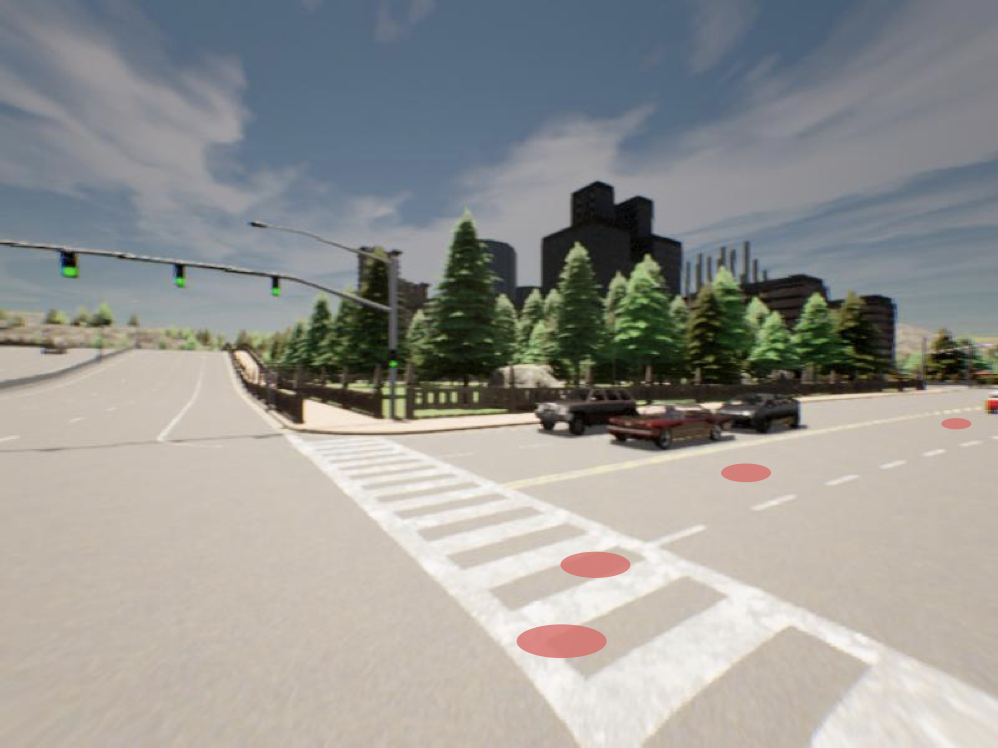}};
            \node[inner sep=0pt, below of=speed-penalty1, yshift=-2cm] (speed-penalty2)
            {\includegraphics[width=5cm]{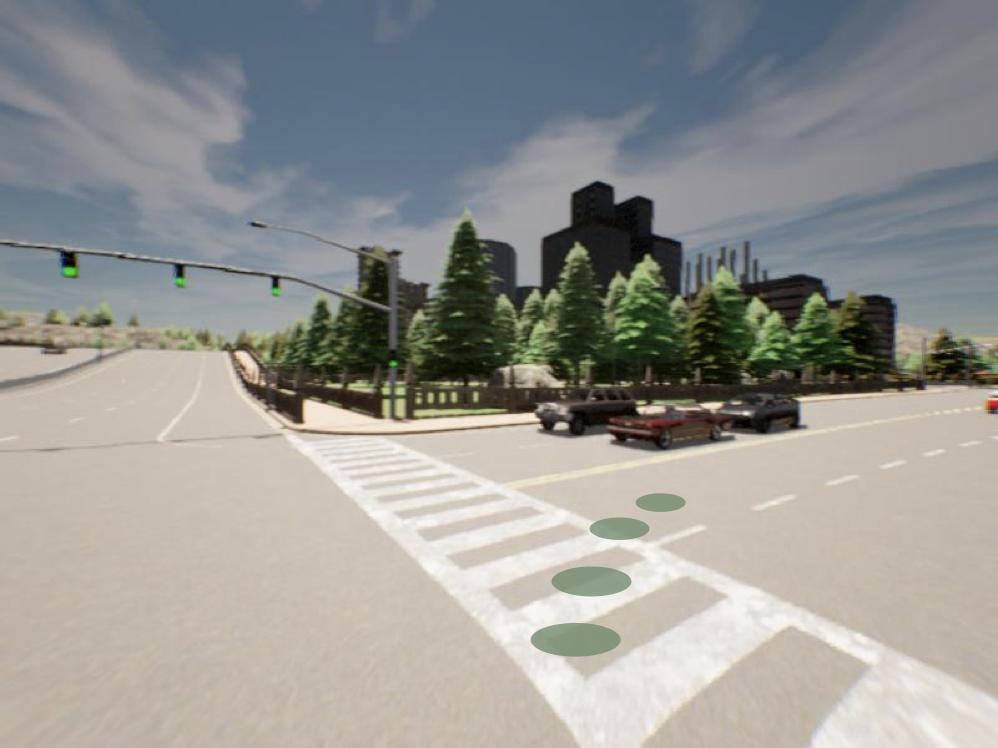}};
            
            \node (legend1) [legend, below of=red-light-penalty2, xshift=-2.25cm, yshift=-0.5cm, minimum width=0.5cm, minimum height=0.5cm, fill=red, label={[label distance=0cm]0: Penalised waypoints}]  {};
            \node (legend2) [legend, right of=legend1, xshift=1.75cm, yshift=0cm, minimum width=0.5cm, minimum height=0.5cm, fill=green, label={[label distance=0cm]0: Compliant waypoints}]  {};
        \end{tikzpicture}
    \end{adjustbox}
    \caption{\textbf{Penalty Illustration.} To ensure compliance with red light and stop penalty rules, as well as promoting deceleration during turning maneuvers, our approach incorporates three distinct penalty types. The first column of the figures exemplifies the red light penalty, wherein waypoints situated beyond the stop line receive a penalty when the traffic light is red. In the second column, we demonstrate the stop sign penalty, wherein predicted waypoints within the vicinity of a stop sign are penalized if the agent fails to decelerate adequately. The speed penalty is enforced during turning actions as shown in last two figures. Specifically, if the predicted waypoints indicate an excessive speed, a speed penalty is imposed. }
    \label{fig:penalties}
\end{figure*}
We found that the objective function design for imitation learning and the autonomous driving metric are not unified which means a low loss of the objective function does not guarantee a high driving score and high route completion. After careful study, we figure out there exist two potential reasons.

\begin{itemize}
    \item The expert agent still makes mistakes when generating the dataset. Sometimes, the expert agent runs a red light and violates the stop sign rule.
    \item The objective function is not sensitive to serious violations of the traffic rules. The average objective function loss may not increase too much when violating the traffic rules despite that this violation may cause serious consequences which result in a huge drop in driving score and route completion.
\end{itemize}

Behavior Cloning (BC), as an imitation learning method, aims to clone the behavior of the expert agent. In such a way, the performance of the trained agent can no longer be better than the expert agent. If the expert agent makes a mistake, the trained agent will learn how to make that mistake instead of getting rid of that mistake.

Our aim is to reformulate the objective function of imitation learning in line with traffic rules, whereby the agent is penalized (higher loss) when it generates short-term future waypoints that violate the traffic rules during the training process.

The traffic rules can be modeled as constrained functions which refer to conditions of the optimization problem that the solution must satisfy. In our setting, we concentrate on two kinds of traffic rule violations and one common driving experience, namely red light violations, stop sign violations, and slowing down when turning, because these are the main problems we found in our vanilla imitation learning approach. We first define three corresponding penalties to quantify these violations. Figure \ref{fig:penalties} illustrates these penalties. 

\subsubsection{Red Light Penalty}
For the red light violation, we design a red light penalty as follows:
\begin{equation}
    {\mathcal{P}}_{\rm tl} = \mathbb{E}_{\mathcal{X}\sim \mathcal{D}}[\mathbb{1}_{\rm red}\cdot\sum_{i=1}^{t}c_i\cdot {\rm max}\{0, w_{i} - \overline{p}\} ]
\end{equation}
where $w_{i}$ is the $i$-th predicted waypoints of the trained agent; $\overline{p}$ is the position of the stop line at the intersection. Both $w_{i}, \overline{p}$ are in the coordinate system of the ego car. $c_i$ is the weight parameter and $\sum_{i} c_i =1$.$\mathbb{1}_{\rm red}$ indicates if a red light that influences the agent exists in the current frame. $\mathcal{X}$ is the input of the current frame and $\mathcal{D}$ is the whole data set. 

In the scenarios of red lights, an extra red light penalty is defined by the distances of the predicted waypoints beyond the stop line at the intersection. If the predicted waypoints are within the stop line, then the penalty remains zero. On the other hand, if the predicted waypoints are beyond the stop line, the sum of distances between those waypoints and the stop line will be calculated as the red light penalty. The additional information for the red light penalty calculation like traffic light information and stop line location is pre-processed and saved in each frame of our dataset.

\subsubsection{Stop Sign Penalty}
Similar to the red light penalty, a stop sign penalty is given when the predicted waypoints violate the stop sign rule. The penalty is formalized as follows:

\begin{equation}
    {\mathcal{P}}_{\rm ss}=\mathbb{E}_{\mathcal{X} \sim \mathcal{D}}[\mathbb{1}_{\rm stopsign}\cdot  {\rm max}\{v - \epsilon, 0\}]
\end{equation}
where $v$ is the desired speed calculated by 
\begin{equation}
    v = \frac{w_{0} - w_{1}}{\Delta t}
    \label{eq: desired speed}
\end{equation}
$w_{0}$ and $w_{1}$ is the first and second predicted waypoint, and $\Delta t$ is the time interval between each frame when collecting the data. $\mathbb{1}_{\rm stopsign}$ is an indicator for stop sign checking. If the vehicle drives into the area that a stop influences, this indicator turns to 1 otherwise it remains zero. $\epsilon$ is the maximum speed required to pass stop sign tests. 

\subsubsection{Speed Penalty}
A speed penalty will be applied if the agent attempts to turn at excessive speed. The motivation to introduce this speed is based on the common driving experience of human beings. Also, we observe the agent sometimes can not avoid hitting pedestrians when turning at high speed since it has less time to react. The penalty is formalized as:
\begin{equation}
    {\mathcal{P}}_{\rm sp}=\mathbb{E}_{\mathcal{X} \sim \mathcal{D}} [{\rm sin}(d\theta) \cdot {\rm max} \{v - v_{\rm lb}, 0\}]
\end{equation}
where $d\theta$ is the direction deviation between the current frame and the next frame. Like stop sign penalty, $v$ is defined in \eqref{eq: desired speed}. $v_{\rm lb}$ is the speed lower bound. Speed under the lower bound will not be imposed by speed punishment. 


With the help of these penalties, the constrained optimization can be formalized as:

\begin{equation}
\label{eq:constraint objective function}
\begin{aligned}
\min \quad & \mathcal{F} \\
\textrm{s.t.} \quad & \mathcal{P}_{\rm tl}, \mathcal{P}_{\rm ss},\mathcal{P}_{\rm sp} = 0\\
\end{aligned}
\end{equation}
where $\mathcal{F}$ is the objective function defined in \eqref{eq:new-designed objective}.

The Lagrange multiplier strategy can be applied here. We introduce three Lagrange Multiplier $\lambda_1$, $\lambda_2$, $\lambda_3$ and the Lagrange function is defined by:
\begin{equation}
    \begin{aligned}
    \min \quad \mathcal{F} + \lambda_1 \mathcal{P}_{\rm tl}+ \lambda_2 \mathcal{P}_{\rm ss} +\lambda_3 \mathcal{P}_{\rm sp}
    \end{aligned}
\end{equation}
This is the final objective function to optimize. For simplicity, these Lagrange multipliers $\lambda_1$, $\lambda_2$, $\lambda_3$ are considered fixed hyper-parameters. Well-chosen $\lambda_1$, $\lambda_2$, $\lambda_3$ are important for optimization. According to our experiments, too large $\lambda$ influences the behaviors in other scenarios while too smaller $\lambda$ is not powerful enough for the agent to obey the corresponding traffic rules. 

The red light indicator and stop sign indicator headers are important for the agent to learn from the stop sign and red light penalty because the information flow of the stop sign and red light helps the agent to build the logistic connection between behavior, observation, and punishment. 
\section{Experiments}
\begin{table*}
    \begin{center}
        \caption{}
        \label{table: Experiment_Long}
        \resizebox{\linewidth}{!}{
        \begin{tabular}{ c | c  c  c | c  c  c  c  c  } 
             \hline
                Model & \makecell[c]{Driving \\ Score \\ \%, $\uparrow$} & \makecell[c]{Route \\ Complication \\ \%, $\uparrow$} & \makecell[c]{Infraction \\ Score \\ $\uparrow$} &  \makecell[c]{Collision \\ Pedestrian \\ \#/km, $ \downarrow$}  & \makecell[c]{Collision \\ Vehicle \\ \#/km, $ \downarrow$} & \makecell[c]{Collision \\ Static \\ \#/km, $ \downarrow$} & \makecell[c]{Red light \\ Infraction \\ \#/km, $ \downarrow$} & \makecell[c]{Stop Sign Infraction \\ Infraction \\ \#/km, $ \downarrow$}\\
            \hline
            \hline
                P-CSG (ours) & $\textbf{56.38} \pm 4.18$ & $\textbf{94.00} \pm 1.75$ & $\textbf{0.61} \pm 0.05$ & $\textbf{0.00} \pm 0.00$ & $\textbf{0.08} \pm 0.02$ & $\textbf{0.00} \pm 0.00$ & $0.03 \pm 0.01$ & $0.01 \pm 0.01$ \\
                TransFuser \cite{TransFuser} & $34.50 \pm 2.54$ & $61.16 \pm 4.75$ & $0.56 \pm 0.06$ & $0.01 \pm 0.01$ & $0.58 \pm 0.07$ & $0.38 \pm 0.05$ & $0.12 \pm 0.03$ & $0.05 \pm 0.02$\\
                TransFuser+ \cite{TransFuser+}& $36.19 \pm 0.90$ & $70.13 \pm 6.80$ & $0.51 \pm 0.03$ & $\textbf{0.00} \pm 0.00$ & $0.40 \pm 0.13$ & $0.04 \pm 0.03$&$0.11 \pm 0.10$ & $0.04 \pm 0.01$ \\
                InterFuser \cite{shao2022interfuser} & $50.64 \pm 3.51 $ & $89.13 \pm 4.12$ & $0.57 \pm 0.05$  & $\textbf{0.00} \pm 0.00$ & $0.09 \pm 0.04$ & $0.01 \pm 0.01$ & $\textbf{0.02} \pm 0.01$ & $0.04 \pm 0.01$\\
                Geometric Fusion \cite{TransFuser} & $31.30 \pm 5.2$  & $57.17 \pm 11.6$ & $0.54 \pm 0.04$ & $0.01 \pm 0.01$ & $0.43 \pm 0.08$ & $0.02 \pm 0.01$ & $0.11 \pm 0.02$ & $0.08 \pm 0.04$\\
                LateFusion \cite{latefusion}& $32.43 \pm 6.72$ & $60.41 \pm 4.23$ & $\textbf{0.61}\pm 0.06 $ & $0.03 \pm 0.02$ & $0.12 \pm 0.03$ & $0.02 \pm 0.01$ & $0.06 \pm 0.02$ & $0.06 \pm 0.02$ \\

             \hline
        \end{tabular}}
        \vspace{1mm}
        \begin{tablenotes}
            \footnotesize
            \item \textbf{Driving Performance.} We evaluate the Driving Score, Route Complication, Infraction Score, collisions, red light violations, and stop sign violations in the Town05 long benchmark. These evaluation metrics are defined by carla leaderboard \cite{Dosovitskiy17}. Note that all other baselines are retrained and tested in our local CARLA environment. Each model is evaluated three times with random seeds. 
        \end{tablenotes}
    \end{center}
    
\end{table*}

\begin{table*}
    \begin{center}
        \caption{}
        \label{table: Ablation Study}
        \resizebox{\linewidth}{!}{
        \begin{tabular}{ c | c  c  c | c  c  c  c  c } 
             \hline
                Model & \makecell[c]{Driving \\ Score \\ \%, $\uparrow$} & \makecell[c]{Route \\ Complication \\ \%, $\uparrow$} & \makecell[c]{Infraction \\ Score \\ $\uparrow$} &  \makecell[c]{Collision \\ Pedestrian \\ \#/km, $ \downarrow$}  & \makecell[c]{Collision \\ Vehicle \\ \#/km, $ \downarrow$} & \makecell[c]{Collision \\ Static \\ \#/km, $ \downarrow$} & \makecell[c]{Red light \\ Infraction \\ \#/km, $ \downarrow$} & \makecell[c]{Stop Sign Infraction \\ Infraction \\ \#/km, $ \downarrow$}\\
            \hline
            \hline
                P-CSG (ours) & $\textbf{56.38} \pm 4.18$ & $\textbf{94.00} \pm 1.75$ & $0.61 \pm 0.05$ & $\textbf{0.00} \pm 0.00$ & $\textbf{0.08} \pm 0.02$ & $\textbf{0.00} \pm 0.00$ & $\textbf{0.03} \pm 0.01$ & $\textbf{0.01} \pm 0.01$ \\
                No CSG & $42.77 \pm 7.78$ & $83.23 \pm 1.86$ & $0.51 \pm 0.11$ & $0.03 \pm 0.01$ & $0.15 \pm 0.04$ & $0.04 \pm 0.03$ & $0.04 \pm 0.00 $ & $0.02 \pm 0.00$\\
                No penalty & $34.98 \pm 5.64$ & $76.20 \pm 13.43$ & $0.51 \pm 0.18 $ & $0.01 \pm 0.00$ & $0.15 \pm 0.11$ & $0.03 \pm 0.01$ & $0.05 \pm 0.02$ & $0.05 \pm 0.01$ \\
                $\lambda_1 = 0.3$ & $37.19 \pm 8.87$ & $82.40 \pm 1.60$ & $0.48 \pm 0.09$ & $0.01 \pm 0.00$ & $0.11 \pm 0.03$ & $0.04 \pm 0.01$ & $0.06 \pm 0.01$ & $\textbf{0.01} \pm 0.00$\\
                $\lambda_1 = 0.7$ & $45.02 \pm 5.54$ & $69.35 \pm 1.90$ & $\textbf{0.70} \pm 0.07$ & $0.01 \pm 0.01$ & $0.09 \pm 0.07$ & $\textbf{0.00} \pm 0.00$ & $0.04 \pm 0.01$ & $\textbf{0.01} \pm 0.00$\\
                $\lambda_2 = 0.005$ & $53.20 \pm 7.02$ & $87.55 \pm 5.73$ & $0.62 \pm 0.10 $ & $0.03 \pm 0.01$ & $0.07 \pm 0.01$ & $\textbf{0.00} \pm 0.00$ & $0.05 \pm 0.02$ & $0.02 \pm 0.02$\\
                $\lambda_2 = 0.05$ & $49.43 \pm 6.73$ & $93.87 \pm 5.51$ & $0.53 \pm 0.02$ & $0.01 \pm 0.01$ & $0.09 \pm 0.04$ & $\textbf{0.00} \pm 0.00$ & $0.05 \pm 0.02$ & $\textbf{0.01} \pm 0.00 $ \\
                $\lambda_3 = 0.3$  & $47.30 \pm 3.58$ & $92.21 \pm 4.52$ & $0.51 \pm 0.07$ & $\textbf{0.00} \pm 0.00$ & $0.11 \pm 0.02$ & $\textbf{0.00} \pm 0.00$ & $0.05 \pm 0.02$ & $0.03 \pm 0.01$ \\
                $\lambda_3 = 0.7$  & $47.35 \pm 4.97$ & $94.45 \pm 3.95$ & $0.50 \pm 0.06$ & $0.01 \pm 0.01$ & $0.13 \pm 0.03$ & $\textbf{0.00} \pm 0.00$ & $0.05 \pm 0.00$ & $\textbf{0.01} \pm 0.01$ \\
             \hline
        \end{tabular}}
        \vspace{1mm}
        \begin{tablenotes}
            \footnotesize
            \item \textbf{Ablation study.} The ablation study for important hyper-parameters. $\lambda_1, \lambda_2, \lambda_3$ are the penalty weight for the red light, speed, and stop sign respectively. The model with no cross-semantics generation structure (No CSG) and no penalty are also given for comparison. Each model is evaluated three times with random seeds. The default weights $\lambda_1$, $\lambda_2$, and $\lambda_3$ we choose for our best model are 0.5, 0.01, and 0.5 respectively. In the process of ablation study, we only change one hyper-parameter and keep others unchanged. 
        \end{tablenotes}
    \end{center}
    
\end{table*}
In this section, our experiment setup will first be described. Then we compare our model against other baselines. We also provide ablation studies to show the improvements from penalty-based imitation learning and cross semantics generation. 

\subsection{Task Description}
The task we concentrate on is a navigation task along the predefined routes in different scenarios and areas. There exists GPS signals guiding the vehicle. Low signal or no signal situations are not taken into consideration. Some predefined scenarios will appear in each route to test the agent's ability to avoid the emergencies, such as obstacle avoidance, other vehicles running a red light, and the sudden appearance of pedestrians on the road. There exists a time limit for the agent to complete the route. Time exceeding is considered a failure in terms of route completion. 
\subsection{Training Dataset}
Realistic driving data is hard to achieve. Alternatively, we use the Carla simulator \cite{Dosovitskiy17} to collect the training data processed by the expert policy. We use the same training dataset as TransFuser \cite{TransFuser}. It includes 8 towns and around 2500 routes through junctions with an average length of 100m and about 1000 routes along curved highways with an average length of 400m. We used the expert agent same as TransFuser to generate these training data.

\subsection{Test Result}
\subsubsection{Benchmark}
We use Town05 long benchmarks to evaluate our model. Town05 long benchmark contains 10 routes and all of these routes are over 2.5km. This benchmark is also used by InterFuser \cite{shao2022interfuser} and TransFuser.
\subsubsection{Baseline}
The other baselines we chose to compare with our model are TransFuser+, TransFuser, Geometric Fusion, and LateFusion. \textbf{TransFuser} \cite{TransFuser} introduces the Transformer into the multi-sensor fusion architecture to achieve better end-to-end autonomous driving results. \textbf{TransFuser+} \cite{TransFuser+}, as an extension of TransFuser, leverages several auxiliary losses to ensure important information flows such as traffic light and road line information in the network. \textbf{InterFuser} \cite{shao2022interfuser} developed a safety control module to regulate the behaviors of the agent, preventing the agent violate the traffic rules. \textbf{LateFusion} \cite{latefusion} uses a simple Multi-Layer Perception Network to integrate multi-modal information. \textbf{Geometric Fusion} \cite{TransFuser} implements both LiDAR-to-image and Image-to-LiDAR fusion to aggregate the information from LiDAR and image to increase the end-to-end autonomous driving ability. 

As Table \ref{table: Experiment_Long} shows, our model achieves the highest route completion and driving scores among all baselines. Compared to TransFuser, Transfuser+, and LateFusion, our model has a huge increase in driving scores and route complications. InterFuser, the current state-of-the-art model, performs well because its safety module avoids dangerous behavior inferred by the neural networks. However, this structure modularizes the decision-making process and these conflicted acts of the safety module and the neural network may have potential risks. Another disadvantage of modular approaches is that the predefined inputs and outputs of individual sub-systems might not be optimal for the driving task in different scenarios. \cite{DBLP:journals/corr/abs-2003-06404} analyses the end-to-end approaches and modular approaches of autonomous driving in detail. 
In contrast to InterFuser, we intend to restrict the behaviors of the agent by introducing penalties to the objective function so that the whole autonomous driving process remains end-to-end. As the results demonstrate, our penalty-based imitation learning can also avoid dangerous behaviors of the agent and make the agent more sensitive to the traffic rules. It achieves even better performance than InterFuser. 

\subsection{Ablation Study}
In this subsection, we will analyze the influences of different penalty weights for corresponding traffic rules. As Table \ref{table: Ablation Study} demonstrates, two extra weights for each penalty are selected for comparison. We also provide the result of models without CSG and penalties for comprehensive analysis. 
We notice that the infractions of traffic lights and stop signs are largely reduced by adding penalties. Our proposed multi-sensor fusion technology (CSG) also decreases the possibility of hitting obstacles such as vehicles, pedestrians, and other statics. 
The results of different penalty weights are also listed in the table for comparison. We found that assigning greater weight to more severe violations will increase the performance of our model. For instance, we apply greater penalties for the red light and the stop sign violations compared to overspeeding by turning, since those two violations cause more serious consequences.

\subsection{Inference and Training Efficiency}
 In this subsection, we aim to compare training time, inference time and parameter number across three SOTA models in the field of end-to-end autonomous driving to gain insights into their computational characteristics. As illustrated in Table \ref{table: efficiency}, our model stands out by having the fewest parameters and the shortest training and inference times when compared to the other two state-of-the-art models. These findings provide compelling evidence of our model's reduced complexity. With fewer parameters and faster processing times, our model showcases an efficient  design, delivering comparable performance while minimizing computational demands. With the shorter inference time, our model also guarantee a safer and more efficient navigation, since the autonomous driving system can quickly detect and react to changes in the environment.
 \begin{table}[htbp]
     \centering
     \caption{}
     \resizebox{\linewidth}{!}{
     \begin{tabular}{c|c c c c}
        \hline
        Model &  \makecell[c]{Parameters \\ Number (M)} & \makecell[c]{Total \\ Training \\ Frames (K)} & \makecell[c]{Training \\ Time  \\ (min. / epoch) } & \makecell[c]{Inference \\ Time  \\ (s / frame)}\\
        \hline
        P-CSG(ours)&  36 & 209 & 33 & 0.043\\
        Interfuser&  53 & 232 & 112 & 0.312\\
        Transfuser+&  168 & 164 & 199 & 0.071\\
        \hline
     \end{tabular}
    }
     \label{table: efficiency}
    \vspace{1mm}
    \begin{tablenotes}
        \footnotesize
        \item \textbf{Complexity.} All experiments are tested with one NVIDIA RTX 6000 GPU and 30 process units under Ubuntu 20.04.6 LTS (Focal Fossa) operating system. The training time per epoch is derived by computing the mean duration across a set of 20 consecutive epochs. The inference time per frame is determined by averaging the elapsed time for 100 individual frames.
    \end{tablenotes}
 \end{table}
\section{Discussion}
In this work, we introduce novel approaches for both Multi-sensor Fusion and Imitation Learning objective function design. The Cross Semantic Generation approach aims to extract and enhance the shared semantic information from LiDAR and RGB inputs. We used some auxiliary losses to regularize the feature space, ensuring the information flow of the features which are important for driving decisions according to human experience. Penalty-based Imitation Learning further increases the level of compliance of the agent with traffic rules. Some other approaches use an extra module to ensure the agent obeys traffic rules.  NEAT \cite{Chitta2021ICCV(NEAT)}, LAV \cite{chen2022lav} introduce some low-level control strategies in the PID controller to force braking at red lights. InterFuser uses a safety module to avoid dangerous actions such as collisions with other vehicles. These strategies largely increase the performance of the agent. However, these extra modules also make the network no longer end-to-end which may potentially cause more tuning effort and suboptimal solutions \cite{DBLP:journals/corr/abs-2003-06404}. With penalty-based Imitation Learning, we aim to avoid those decisions detached from the network. We use the penalty to make the agent more sensitive to traffic rules. The end-to-end nature of the network is guaranteed while constraining the agent to comply with traffic regulations. 



Our study also has several limitations. First of all, we only use front-view images and 180-degree LiDAR data as input. The information from the rear of the vehicle is missing which may cause collisions when changing the lane. Besides, we only tried to integrate the RGB and LiDAR input, more sensors like radar input and depth-map can be taken into consideration. Additionally, we only test the performance of our model in the simulation environment. Real-world data can be more complex and contain more noise. Finally, we believe that further research should be dedicated to introducing additional penalties into the objective functions, as this approach holds promise in the development of human-level end-to-end autonomous driving agents.
\section{Conclusion}
The key points for the performance of end-to-end autonomous driving are improving fusion technologies and policy learning methods. These two points turn into two important questions. How to efficiently extract and integrate the features from different modalities? How to effectively use these features to learn a stable and well-performing policy approaching or even surpassing the human level. In this paper,  we contribute to the abovementioned aspects and achieve state-of-the-art performance. Compared to modular autonomous driving technologies, end-to-end autonomous driving has lower hardware costs and less expensive maintenance. It is also adaptable to different scenarios simply by feeding data. We believe end-to-end autonomous driving can be deployed in actual vehicles in the near future.

\section{Acknowledgement}
This work is supported by Huawei Trustworthy Technology and Engineering Laboratory.  We thank Prof. Fengxiang Ge and Wei Cao for the insightful discussion.

{
\small
\bibliographystyle{IEEEtran}
\bibliography{reference}
}

\end{document}